\pdfoutput=1

\documentclass[11pt]{article}

\usepackage{acl}

\usepackage{times}
\usepackage{latexsym}

\usepackage[utf8]{inputenc} 
\usepackage[T1]{fontenc}    
\usepackage{hyperref}       
\usepackage{url}            
\usepackage{booktabs}       
\usepackage{amsfonts}       
\usepackage{nicefrac}       
\usepackage{microtype}      
\usepackage{lipsum}		
\usepackage{multirow}
\usepackage{graphicx}
\usepackage{natbib}
\usepackage{doi}
\usepackage{wrapfig}
\usepackage[ruled,vlined]{algorithm2e}
\usepackage{enumitem}
\usepackage{array}

\usepackage{amsmath}
\usepackage{amssymb}
\usepackage{mathtools}
\usepackage{amsthm}
\usepackage{xcolor}
\usepackage[most]{tcolorbox}
\definecolor{ForestGreen}{RGB}{34,139,34}
\usepackage{cleveref}

\theoremstyle{plain}
\newtheorem{theorem}{Theorem}[section]

\newtheorem{corollary}[theorem]{Corollary}
\theoremstyle{definition}
\newtheorem{definition}[theorem]{Definition}

\title{Speculative Diffusion Decoding:\\Accelerating Language Generation through Diffusion}

\date{} 					
\author{
Jacob K.~Christopher\\
University of Virginia\\
\texttt{csk4sr@virginia.edu}
\And
Brian R. Bartoldson \\
Lawrence Livermore National Laboratory \\
\texttt{bartoldson1@llnl.gov}
\AND 
Tal Ben-Nun \\
Lawrence Livermore National Laboratory \\
\texttt{talbn@llnl.gov}
\And
Michael Cardei \\
University of Virginia \\
\texttt{ntr2rm@virginia.edu}
\AND
Bhavya Kailkhura \\
Lawrence Livermore National Laboratory \\
\texttt{kailkhura1@llnl.gov}
\And
Ferdinando Fioretto \\
University of Virginia \\
\texttt{fioretto@virginia.edu}
}

\hypersetup{
pdftitle={Speculative Diffusion Decoding},
pdfsubject={cs.CL, cs.LG},
pdfkeywords={Parallel Decoding, Large Language Models, Discrete Diffusion Models},
}

\newcommand{\nando}[1]{{\textcolor{blue}{Nando: \footnotesize\sf[#1]}}}

\begin{document}
\maketitle
\begin{abstract}
Speculative decoding has emerged as a widely adopted method to accelerate large language model inference without sacrificing the quality of the model outputs. 
While this technique has facilitated notable speed improvements by enabling parallel sequence verification, its efficiency remains inherently limited by the reliance on incremental token generation in existing draft models. To overcome this limitation, this paper proposes an adaptation of speculative decoding which uses discrete diffusion models to generate draft sequences. This allows parallelization of both the drafting and verification steps, providing significant speedups to the inference process. Our proposed approach, {\it Speculative Diffusion Decoding (SpecDiff)}, is validated on standard language generation benchmarks and empirically demonstrated to provide up to 7.2x speedups over standard generation processes and up to 1.75x speedups over existing speculative decoding approaches.
\end{abstract}

\begin{figure*}[t]
\centering
\includegraphics[width=0.99\linewidth]{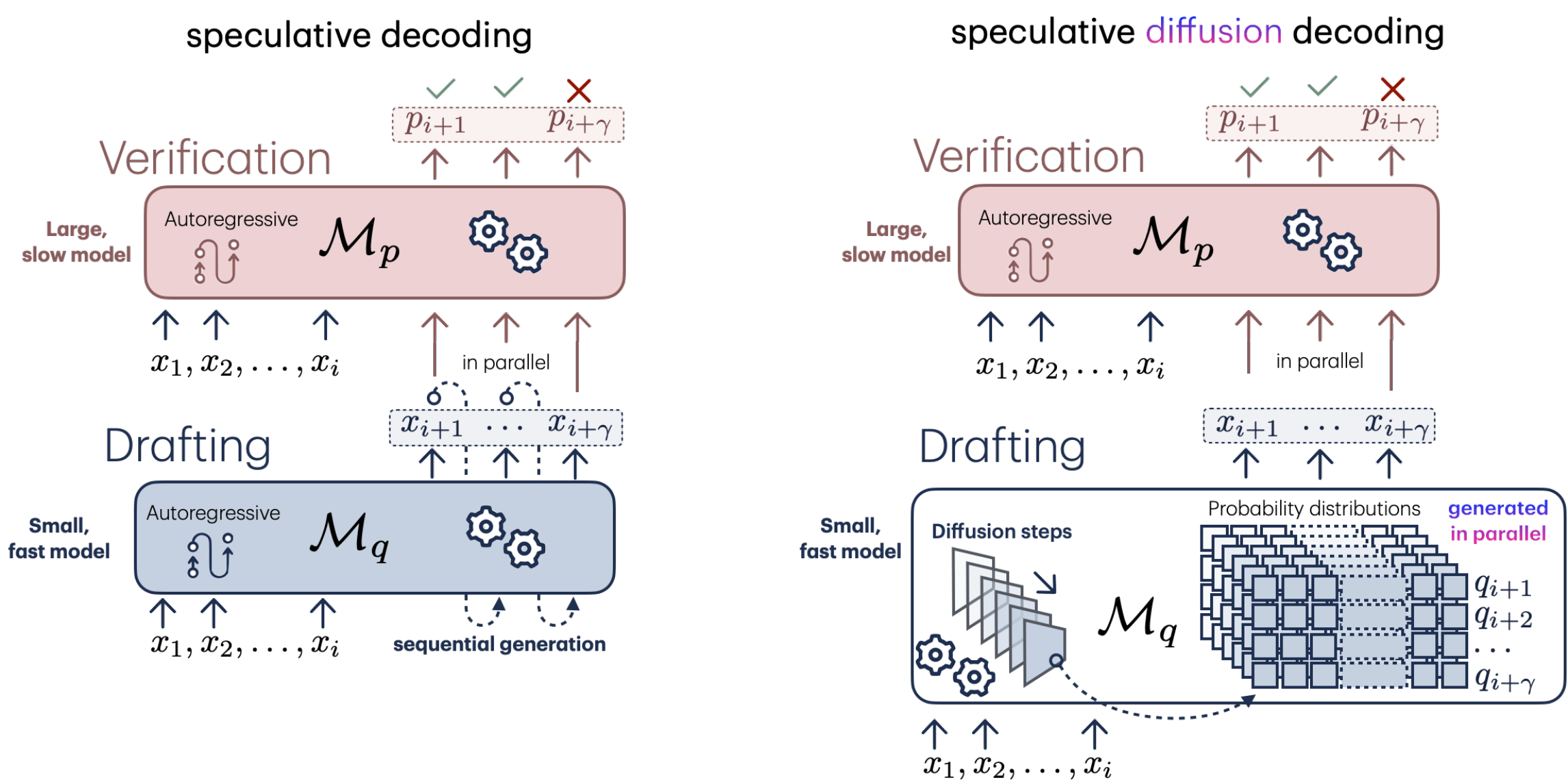}
\caption{Illustration of classical speculative decoding (left) and 
speculative diffusion decoding (right).}
\label{img:scheme}
\end{figure*}

\section{Introduction}
\label{sec:introduction}

As autoregressive language modeling with transformers \citep{vaswani2017attention} is scaled to larger compute levels, performance improves and new capabilities emerge \citep{ kaplan2020scaling, brown2020language}. Indeed, scaling has been shown to to improve the performance of large language models (LLMs) for a diverse array of tasks, including code generation, question answering, summarization, and many other use cases \citep{achiam2023gpt,team2023gemini,team2024llama}. 
For instance, models such as 
LLaMA 3.2 90B \citep{team2024llama}, ChatGPT \citep{openai2024gpt4technicalreport}, Cohere 52B \citep{cohere}, Google’s Gemini-ULTRA \citep{geminiteam2024} exemplify the ongoing trend of deploying and releasing increasingly large models, enabling broader access and application across various domains. 

However, while these desired capability arise, running LLMs in inference mode for millions of users produces burdensome electricity, time, and monetary demands. Many methods exist to mitigate these costs -- including sparsity, quantization, and distillation -- but they often introduce new tradeoffs, e.g., their application can degrade the performance of the model~\citep{hongdecoding}.

Unlike other methods for accelerating LLM inference, \emph{speculative decoding} \citep{xia2023speculative, leviathan2023fast} can improve LLM efficiency by $2\mbox{--}3\times$ with \emph{no degradation in the quality of the model outputs}. In \citet{leviathan2023fast}, speculative decoding achieves this by sequentially generating multiple tokens with a small, efficient draft model, then running the target, large, LLM in parallel on all of the drafted tokens, simultaneously evaluating their consistency with the target LLM's output token probabilities. 
Provided that the drafting model’s tokens are frequently accepted by the target model and that the drafting model operates substantially faster than the target model, speculative decoding can directly match the sampling output from the target model while significantly reducing runtime  \citep{leviathan2023fast}. This functioning is shown in Figure~\ref{img:scheme} (left).

Notably, since both the drafting model's speed and its alignment relative to the target model are critical to the success of speculative decoding, simultaneous improvements in each of these areas are necessary to ensure speculative decoding's relevance to future, more capable target models. For instance, a small GPT-2 \citep{radford2019language} drafting model could produce drafts that are often rejected by GPT-4 \citep{achiam2023gpt}, and simply scaling the drafting model to address its weaker generations risks diminishing the speed advantage necessary to speculative decoding's success.
To address this challenge, previous efforts have focused on 
introducing additional parallelization techniques that incorporate prediction trees and branching to refine the drafting process \citep{fu2024break,miao2023specinfer,svirschevski2024specexec}. 
However, the gain in generation efficiency are at the expense of much increased number of operations and/or memory for each generation.

This paper proposes a fundamentally different approach to improve speculative decoding: \emph{It proposes to replace the auto-regressive drafter with recently introduced discrete diffusion models} \citep{loudiscrete,sahoo2024simple}. These models offer several key advantages when used as drafters: Firstly, they provide a smooth trade off between the compute cost of generation and the quality of generation (via the number of reverse diffusion steps). Second, while they have historically struggled relative to traditional language models, recent diffusion models have been shown to require $32\times$ fewer function evaluations than autoregressive models to produce text with comparable perplexity \citep{loudiscrete}, with more recent works reporting even further speedups \citep{sahoo2024simple}. This is a trend which is anticipated to continue. 
Finally, future advances in diffusion model generation quality are highly aligned with their ability to perform strongly as speculative drafters: as drafted tokens are accepted by the target model at a higher rate, a larger number of proposed drafted tokens becomes optimal from an efficiency/speed point of view, and (unlike sequential drafters) diffusion models can easily accommodate generation of many more tokens since they are able to generate entire sequences in a single step. 

\paragraph{Contributions.}
More specifically, this paper makes the following contributions: 
\begin{enumerate}[leftmargin=*, parsep=2pt, itemsep=2pt, topsep=0pt]
\item  It introduces a novel integration of generative diffusion language models with speculative decoding, schematically illustrated in Figure \ref{img:scheme} (right).
\item  It empirically demonstrate the hybrid model’s ability to significantly accelerate inference times while maintaining the same high-quality outputs of the original, target large language model.
\item  The proposed method ensures that all generations from the diffusion language model, which are empirically shown to produce outputs with significantly higher perplexity than current state-of-the-art autoregressive models \citep{sahoo2024simple, loudiscrete, austin2021structured, gloeckle2024betterfasterlarge}, align with the outputs generated by larger, more computationally demanding models.
\item  Finally, the paper sets a new benchmark for speed in language completion tasks on the CNN/DM  \cite{nallapati2016abstractive}and OpenWebText datasets \cite{gokaslan2019openwebtext}. 
\end{enumerate}

\section{Related Work}
While autoregressive language models provide state-of-the-art performance on language generation tasks, the incremental decoding used by these architectures results in significant overhead at inference time \citep{miao2023towards}. This is largely a result of the inability to parallelize the sequential process of generating tokens in the output sequence as each token generation is dependent upon the preceding tokens in the sequence; consequentially, scaling the compute associated with the inference cannot directly reduce this overhead when using standard decoding schemes. In recent literature studying how to accelerate large language model generation, two primary approaches have been explored: (1) advanced decoding implementations that better parallelize token generation and (2) non-autoregressive language models allowing full sequences to be generated simultaneously.

\paragraph{Speculative decoding.}
Speculative decoding accelerates autoregressive generation by leveraging a smaller autoregressive models of the same architecture (the drafter model) to predict candidate sequences, which the original model (the target model) then verifies \citep{leviathan2023fast,chen2023accelerating}.
Notably, the earliest literature on speculative diffusion adapted a non-autoregressive model to act as the drafter model \citep{xia2023speculative}, using a masked language model with a bidirectional decoder \citep{ghazvininejad2019mask}.
However, the integration of non-autoregressive draft models has not received much attention due to the difficulty introduced by the necessary additional training in existing approaches and the modest speedups that were previously reported using these methods (less than 2x speedup over vanilla decoding schemes).

Thus, recent advancements in speculative decoding have focused on overcoming memory-related constraints, with improvements achieved through various approaches: drafting directly with the target model \citep{cai2024medusasimplellminference,zhang2024recurrentdrafterfastspeculative}, enhancing draft algorithms \citep{sun2024spectrfastspeculativedecoding}, and introducing additional parallelization techniques that incorporate branching to refine the drafting process \citep{fu2024break,miao2023specinfer,svirschevski2024specexec}. 

\paragraph{Non-autoregressive language models.}
Models which stray from the autoregressive paradigm have been shown to speedup generation by generating blocks or even entire sequences simultaneously. \citet{gloeckle2024betterfasterlarge} propose a method of adapting traditional autoregressive models to sample blocks of tokens, improving inference time over similarly scaled models. 
In a similar vein, diffusion language models have been recognized for their efficiency in generating extended token sequences concurrently, offering even greater speed enhancements. 
These models recast language generation as a diffusion process either across the embedding space \citep{austin2021structured} or, more recently, through the probability distributions of generated tokens \citep{sahoo2024simple, loudiscrete}.
Recent models report up to a 32x speedup over similarly sized GPT-2 models \cite{loudiscrete}, and current state-of-the-art further improves runtime speed \cite{sahoo2024simple}.
However, despite they have been shown to dramatically accelerate the inference time for language generation, diffusion models typically perform less effectively than state-of-the-art autoregressive models in terms of standard language metrics, often exhibiting significantly higher perplexity scores \cite{zheng2024masked}. In the following section, {\it we will demonstrate, for the first time, how the speed of these models can be leveraged without being subject to this critical limitation.}

\section{Preliminaries and Settings}
\label{sec:preliminaries}

For open-ended language generation, we focus on the task of token generation, where given a sequence of tokens $x_1, x_2, \ldots, x_i$, 
denoted here with shorthand notation $x_{1:i}$, the goal is to generate the next $n$ tokens $x_{i+1}, \ldots, x_{i+n}$ from the conditional distributions 
$p(x_{i+1} | x_{1:i}), \ldots, p(x_{i+n} | x_{1:i+n-1})$ or more succinctly $p_{i+1}, \ldots, p_{i+n}$. 

\paragraph{Speculative decoding.}
Speculative decoding leverages two LLMs, $M_p$ and $M_q$, to parallelize token generation:
\begin{itemize}[leftmargin=*, parsep=4pt, itemsep=0pt, topsep=4pt]
\item $M_p$ is the original, \emph{target}, model whose output probability distributions for the tokens are $p_{i+1}, \ldots, p_{i+n}$. 
\item $M_q$ is a smaller and more efficient \emph{drafter} model,  used to generate approximations of the distribution of $M_p$ as $q_{i+1}, \ldots, q_{i+n}$. 
\end{itemize}
This process follows a \textit{draft-then-verify} approach \citep{stern2018blockwise}, where $M_q$ efficiently computes a candidate sequence of tokens, which $M_p$ then verifies in parallel.

During each speculative decoding iteration, $M_q$ generates a subset of the total $n$ tokens that are required for the generation task. The size of this subset is denoted as $\gamma$. 
As shown in Figure \ref{img:scheme} (left), the tokens $x_{i+1 : i+\gamma}$ sampled from $M_q$ are then used by $M_p$ to generate the corresponding probability distributions $p_{i+1}, \ldots, p_{i+\gamma}$. 
The distributions $q_{i+1}, \ldots, q_{i+\gamma}$ from $M_q$ are stored for evaluating acceptance in subsequent steps. Critically, the target model's inference over $p_{i+1}, \ldots, p_{i+\gamma}$ can now be run in parallel as the model has access to tokens $x_{i+1 : i+\gamma}$, alleviating the sequential dependency for generation with $M_p$.

To ensure high-quality outputs despite potential discrepancies between $M_p$ and $M_q$, tokens are subjected to an acceptance criterion. For each token $x_j$ with $j \in [i+1, i+\gamma]$, if $q(x_j) \leq p(x_j)$, the token is accepted. If $q(x_j) > p(x_k)$, the token is rejected with a probability of $1 - \frac{p(x_j)}{q(x_j)}$. This criterion is applied sequentially from left to right; rejection of any token results in the discard of all subsequent tokens. Hence, the token acceptance is maximized when the output distributions of $M_q$ and $M_p$ are closely aligned. 

Previous literature quantifies the likelihood of token acceptance, denoted $\alpha$, and theoretically demonstrate that $\alpha = 1 - \mathbb{E}(D_{LK}(p,q))$ where $D_{LK}$ represents a divergence measure between distributions \cite{leviathan2023fast}. This has led to the prevalent use of drafters taken from the same series as the target models, \emph{a paradigm that we challenge in this paper}.

\section{Speculative Diffusion Models}
\label{sec:method}

Speculative decoding has provided state-of-the-art results for improving language generation inference time but requires meticulous tuning of the associated hyperparameters to achieve optimal results. Particularly $\gamma$, the sequence length generated by the drafter model, needs to be appropriately calibrated not only to maximize potential speedup but to even outperform standard autoregressive decoding. This is an important consideration when using current autoregressive draft models, provided that the inference time to generate $M_q(x)$, the draft logits, is directly scaled by the size of $\gamma$. Increasing this value too high reduces the number of operations that are conducted in parallel, {\it potentially leading to speculative decoding increasing inference time,} while reducing this value too low results in speculative decoding ``missing out'' on token generations that could have been handled by the draft model.

\citet{leviathan2023fast} has conducted theoretical analysis on how to best optimize the value of $\gamma$, however, it has been contingent upon accurately estimating the percentage of tokens in a the sequence that will be accepted by the target model. By their own acknowledgment, it would be necessary to predict this value {\it for each draft} and numerically solve for the optimal value of $\gamma$ to fully realize the potential speedup of speculative decoding. Thus, a significant portion of the residual suboptimality in current implementations can be attributed directly to the sensitivity of this hyperparameter.

Diffusion language models are juxtaposed to conventional language models in that they do not sample token sequences in a sequential manner, rather generating entire sequences in parallel. This has resulted in significant speedup over similarly sized autoregressive models when generating extended sequences \citep{loudiscrete}. This can particularly be observed in longer sequence generations as scaling the draft length $\gamma$ results in minimal overhead due to the ability to directly parallelize token generation.



\subsection{SpecDiff: Formulation}


Diffusion models generally operate on continuous data spaces by progressively adding Gaussian noise to the data in a forward diffusion process and then learning to reverse this process to generate new samples \cite{sohl2015deep, ho2020denoising, song2019generative}. This framework is well-suited for data types such as images, where pixel values can be treated as continuous and thus can naturally accommodate additive Gaussian noise. However, when dealing with discrete data like natural language (tokens), the assumption of continuous noise addition does not hold, as it would result in non-integer values that do not correspond to valid tokens. 

Discrete diffusion models address this limitation by redefining the  forward processes to transition between discrete token states, such as replacing tokens according to a transition matrix or introducing randomness through categorical distributions \cite{austin2021structured, hoogeboom2021argmax, sohl2012discrete}. This process enables the generation of coherent and meaningful sequences in natural language processing tasks.
However, unlike continuous diffusion models, traditional score-matching techniques cannot be directly applied to learn discrete diffusion models. Instead, various surrogate losses have been proposed for training. 

In particular, \citeauthor{sahoo2024simple}~introduce the Masked Diffusion Language Model (MDLM), which gradually masks and then reconstructs tokens within a text sequence, enabling efficient text generation. MDLM optimizes a continuous-time Negative ELBO (NELBO)  objective to minimize the negative log-likelihood over a continuous-time diffusion process, which can be formulated as follows:

\begin{equation}
\scriptsize
L_{\infty}^{\text{NELBO}} = \mathbb{E}_{\tilde{q}} \left[ \int_{t=0}^{t=1} \frac{(1-\beta_t)'}{\beta_t} \sum_{i} \log \langle x_{\theta}(z^t_i), x_{i} \rangle \, dt \right]
\end{equation}
where \( x_{\theta}(z^t_i) \) represents the model's estimate of the original token \( x_i \) at time \( t \) given the current noisy state \( z^t_i \), and \( \beta_t \) denotes the noise schedule controlling the diffusion process. Here, $\tilde{q}$ is the forward noising process of in the masked diffusion, defining the distribution over the noisy latent variable $z_i^t$, and can be related to the NELBO as described in Equation (8) of 
\citealp{sahoo2024simple}. The expectation $\mathbb{E}_{\tilde{q}}$ signifies averaging over the possible outcomes of  $\tilde{q}$ allowing the model to accound for all possible variations of $z_t$.
The Noise Schedule Derivative term, $\frac{(1-\beta_t)'}{\beta_t}$, represents the rate of change in the noise schedule \( \beta_t \) over time, and $t$ is a continuous timestep between 0 and 1.

This loss is directly used for pretraining our draft model. As this process learns to denoise over the probability mass vectors, the output of the draft model is a matrix of  $\mathbb{R}^{n \times m}$ where $n = \gamma$ and $m$ is the size of the vocabulary. 
The candidate sequence can then be generated using standard decoding methods over these probability mass vectors, the logits of which are stored to be used when determining whether to accept each draft token.

\newcommand{\COMMENTLLAMA}[1]{\vspace{3pt}{\color{ForestGreen} $\triangleright$ {#1}}}
{
\begin{algorithm}[tb]
\DontPrintSemicolon
\caption{SpecDiff Decoding}
\label{alg:spec-diff}
{\fontsize{9pt}{7.4pt}\selectfont
    \COMMENTLLAMA{Take $T$ diffusion steps to generate the draft.} \\
    $q_{i+1,\ldots,i+\gamma}^T \sim \mathcal{N}(\mathbf{0}, \sigma_T \mathbf{I})$ \\
    \For {$t = T$ \KwTo 1}{
        $q_{i+1,\ldots,i+\gamma}^{t-1}(x) \gets M_q([x_{0 : i}] \circ [q_{i+1,\ldots,i+\gamma}^t(x)], t)\!\!\!\!$
    
    }
    $x_{i+1,...,i+\gamma} \sim {q}^0$ \\
    \COMMENTLLAMA{Run $M_p$ in parallel.} \\
  $p_i(x), \ldots, p_{i + \gamma + 1}(x) \gets$ 
  $M_p(x_{0:i}), \ldots, M_p(x_{0:i+\gamma})$ \\
  \COMMENTLLAMA{Determine the number of accepted guesses $n$.} \\
  $r_{i} \sim U(0, 1), \dots, r_{i+\gamma} \sim U(0, 1)$ \\
  $n \gets \min(\{ j - 1 \mid i \le j \le i+\gamma, r_j > \frac{p_j(x)}{q_j(x)} \} \cup \{ \gamma \})$ \\
  \COMMENTLLAMA{Adjust the distribution from $M_p$ if needed.} \\
  $p'(x) \gets p_{n+1}(x)$ \\
  \If{$n < i+\gamma$}{
    $p'(x) \gets norm(max(0, p_{n+1}(x) - q_{n+1}(x)))$
  }
  \COMMENTLLAMA{Return one token from $M_p$, $n$ tokens from $M_q$.} \\
  $t \sim p'(x)$ \\
  \Return $x_1, \ldots, x_{n}, t$
}
\end{algorithm}
}
Now, the draft logits produced by the output matrix of the discrete diffusion drafter directly substitute the autoregressive drafter used to generate $M_q([x_{0:i}] \circ [x_{i+1}, \ldots, x_{i+\gamma}])$, where $\circ$ is a list concatenation operator. 
This substitutes the draft step taken by~\citeauthor{leviathan2023fast}~and~\citeauthor{chen2023accelerating}~and is the primary difference between SpecDiff and standard speculative decoding approaches. The subsequent steps of verifying this draft with the target model follow the previously proposed decoding algorithm, thus our proposal requires minimal modifications to existing speculative decoding code bases. 

A complete overview of the SpecDiff decoding is provided in Algorithm \ref{alg:spec-diff} (adapted from~\citeauthor{leviathan2023fast}).

\paragraph{Drafter's evaluations.}
Next, we highlight an important difference between standard speculative decoding and our approach. 
While in standard speculative decoding the number of evaluations by the drafter model is dictated by the value of $\gamma$ (used in the first loop for Algorithm \ref{alg:spec-diff}), in speculative diffusion it is dictated by the number of diffusion steps, $T$. This allows SpecDiff to scale $\gamma$ to higher values, as discussed further in Section \ref{subsec:results-discussion}. 

Instead, the value of $T$ is selected to optimize the trade-off between draft quality and computational overhead. While analysis by \citeauthor{loudiscrete}~shows that lower values of $T$ lead to higher perplexity in the generated sequence, this only impacts SpecDiff with respect to its effect on the percentage of tokens from the draft which are accepted (as can be noted in the analysis reported in Figure \ref{fig:hyperparam-analysis} and discussed in details Section \ref{subsec:results-discussion}).

\paragraph{Sequence initialization.}
An important consideration in implementing SpecDiff is the initial alignment between the diffusion draft model and the target model’s data distribution since these models architectures are fundamentally different one another. 
A key strength of SpecDiff lies in its ability to leverage the alignment between the prefixes used in discrete diffusion and the target model’s data distribution. Specifically, the better the prefixes align with the target distribution, the more effectively the diffusion drafter can generate longer, coherent sequences matching the target distribution, resulting in progressively higher acceptance rates. 

This however, also means that when the diffusion draft model has not been finetuned, its output distribution may initially differ from that of the target model, potentially leading to lower acceptance rates at the beginning of the generation process. To address this, we employ standard speculative decoding for the initial few tokens, thereby optimizing SpecDiff’s performance from the outset. This strategy ensures that SpecDiff achieves speedups that are empirically at least as significant as those observed with standard speculative decoding and can realize substantial improvements, as we show in Tables \ref{tab:walltime-metrics-prefix} and \ref{tab:walltime-metrics}. Moreover, once the diffusion draft model is finetuned to better match the target distribution, this initial speculative decoding becomes unnecessary, as the drafter effectively aligns with the target model from the beginning.

\begin{table*}[!t]
        \vspace{0pt}
        \centering
        \resizebox{\linewidth}{!}{
        \begin{tabular}{lccccc} 
            \toprule
            \textbf{Method} & \textbf{Total Memory} & \textbf{FLOPs} & \textbf{Depth} & \textbf{Max}   & \textbf{Extra Training} \\
                            & \textbf{Consumption}    &                &                & \textbf{Tokens} & \textbf{Requirements} \\ 
            \midrule
            Autoregressive     & $C_p(1)$          & $F_p(1)$ & $D_p$ & 1  & N/A           \\ 
            \midrule
            SpS~\cite{leviathan2023fast}\textsuperscript{1} & $C_p(\gamma) + C_q(\gamma)$ & $F_p(\gamma) + F_q(\gamma)$ & $D_p + \gamma D_q$ & $1+\gamma$& $M_q$ \\ 
            PLD~\cite{pld}    & $C_p(\gamma)$ & $F_p(\gamma) + \mathcal{O}(Pn)$ & $D_p + \mathcal{O}(n)$ & $1+\gamma$ & None \\ 
            Lookahead~\cite{fu2024break}\textsuperscript{1} & $C_p\left(2\gamma(n-1)\right)$ & $F_p\left(2\gamma(n-1)\right)+F_p(1)$ & $D_p$ + $\mathcal{O}(1)$ & $n-1$ & None \\ 
            Medusa-1~\cite{cai2024medusasimplellminference}  & $C_p(\mathcal{T}_n) + KC_q(\gamma)$ & $F_p(\mathcal{T}_n) + KF_q(\gamma)$ & $D_p+D_q$ & $1+\mathcal{T}_d$& $M_q$ \\ 
            Medusa-2~\cite{cai2024medusasimplellminference}\textsuperscript{2}   & --- & --- & --- & --- & $M_p$ \\ 
            Hydra~\cite{hydra}\textsuperscript{3}      & --- & --- & $D_p+K D_q$ & ---  & $M_q$ \\ 
            EAGLE~\cite{eagle}     & $C_p(25) + C_q(25)$ & $F_p(25) + 5F_q(5)$ & $D_p + 5D_q$ & $6$ & $M_q$ \\
            EAGLE-2~\cite{eagle2}\textsuperscript{4}   & $C_p(\mathcal{T}_n) + C_q(\mathcal{T}_n)$& $F_p(\mathcal{T}_n) + \mathcal{T}_d F_q(\frac{\mathcal{T}_n}{\mathcal{T}_d})$ & $D_p + \mathcal{T}_d D_q$ & $1+\mathcal{T}_d$ & $M_q$ \\ 
            \midrule
            SpecDiff (Ours) & $C_p(\gamma) + C_q(\gamma)$ & $F_p(\gamma) + T F_q(\gamma)$ & $D_p + T D_q$ & $1+\gamma$& $M_q$ \\ 
            \bottomrule
        \end{tabular}
        }
        \caption{Comparison of different speculative decoding strategies and their parallel properties. The quantities represent the cost of one additional decoding step (i.e., following the initial prompt). Note that $M_q$ (and related quantities such as $F_q$) are not constant across methods; e.g., EAGLE uses a different draft model architecture than SpecDiff. $T$ is the number of diffusion steps, $\mathcal{T}$ is a sparse draft tree with $\mathcal{T}_n$ nodes and depth $\mathcal{T}_d$, $K$ is the number of heads in a multi-head speculative decoder, $P$ is the number of prompt tokens, $n$ is an n-gram length, and $\gamma$ is the number of proposed tokens. \textsuperscript{1} $\mathcal{O}(1)$ represents a database lookup. \textsuperscript{2} Medusa-1 with target fine-tuning for Medusa-2. \textsuperscript{3} Medusa with sequential draft heads for Hydra. \textsuperscript{4} EAGLE with dynamic draft trees for EAGLE-2.}
        \label{tab:baseline-comparison}
\end{table*}

\section{Comparative Analysis of Speculative Decoders}
\label{appendix:specdecs}

\Cref{tab:baseline-comparison} compares the computational aspects of SpecDiff with state-of-the-art baselines in speculative decoding.

According to the Work-Depth parallel computation model~\cite{workdepth}, an algorithm can be represented by a DAG, in which each node is an operation and each edge is a dependency. The longest shortest path in this computational DAG (i.e., the longest dependency chain) is called the \textit{depth}. It follows that the average parallelism of a computation is the total number of nodes divided by the depth, and that higher depth means fewer opportunities for parallelism. We represent the operations by the FLOPs of the model.

The table reasons about the approaches through three parameters for $M_p$ and $M_q$: $C$ (memory consumption), $F$ (arithmetic/floating-point operations, or FLOPs), and $D$ (depth). Due to the quadratic memory and computation requirements of transformers, and for simplicity, $C_{\{p,q\}}$ and $F_{\{p,q\}}$ are represented as functions of the number of tokens after the initial prompt, i.e., $C_p(k)$ is the memory cost for generating $P+k$ tokens, where $P$ is the number of prompt tokens. The depth $D_{\{p,q\}}$ is generally independent of the number of tokens computed and is thus represented by a scalar.

In Medusa, the prediction tree is set by a number of fixed hyperparameters~\cite{cai2024medusasimplellminference}, which we aggregate by using the number of nodes and the depth of the tree $\mathcal{T}_n$ and $\mathcal{T}_d$. Medusa-2 exhibits the same parallel properties as Medusa-1, but requires a joint fine-tuning of $M_p$ along with $M_q$.
Hydra is an adaptation of Medusa, in which the heads are applied in sequence. It thus only affects the depth of the computations.

EAGLE uses a drafter model that contains the target embedding layer, one autoregressive layer, and the target LM head. 
The fixed tree size is 5 levels deep and contains 25 nodes~\cite{eagle}. EAGLE-2~\cite{eagle2} dispenses with the fixed tree defined in EAGLE, and the induced dynamic tree used in generation is represented with $\mathcal{T}$.

Notice that in our method, the depth of the algorithm is dependent on $T$, the number of diffusion steps, rather than $\gamma$. Empirically, $T$ is smaller than $\gamma$, which enables more parallelism in the computation w.r.t. the number of generated tokens. Combined with diffusion models enabling longer prediction horizons, this allows SpecDiff to produce more speculative predictions faster.

Practically, we observe that this enables SpecDiff to produce near state-of-the-art inference acceleration while requiring significantly less computational overhead and reduced memory requirements. 



\section{Evaluation}

\subsection{Experimental Setup}

\paragraph{Settings.}

To empirically evaluate the improvements provided by using SpecDiff, the paper provides an empirical analysis on 
three standard natural language processing tasks: {\bf (1)} text summarization using the CNN/DM dataset \cite{nallapati2016abstractive}, {\bf (2)} text generation on the OpenWebText (OWT) dataset \cite{gokaslan2019openwebtext}, and {\bf (3)} text generation using MT Bench \cite{zheng2023judging}. 
Additionally, we assess the performance of our method against the current state-of-the-art using {\it SpecBench}, a unified evaluation platform for speculative decoding techniques \cite{xia-etal-2024-unlocking}.

In each setting, the model is queried for 1024 tokens using a greedy decoding scheme (temperature = 0). For the experiments, we evaluate the pretrained target models $M_p$ GPT-2 XL (1.5B), GPT-NEO (2.7B), and Vicuna (33B). We use Masked Diffusion Language Model (110M) as our drafter model $M_q$, which is a comparable size to the baseline drafter GPT-2 (86M) employed for our standard speculative decoding baseline \cite{sahoo2024simple}. 

All evaluation is conducted on two NVIDIA A100 series GPUs (80GB) using CUDA 12.2. Additionally, FlashAttention \cite{dao2022flashattention} is used to optimize the performance in all experiments.


\paragraph{Evaluation metrics.}

Our method is assessed empirically by walltime speedup, acceptance rate $\alpha$, and total floating point operations (FLOPs).
The reported results are compared to recognized baselines of vanilla autoregressive decoding, standard speculative decoding (SpS) implementations as proposed by \citeauthor{leviathan2023fast,chen2023accelerating}, which our method most closely resembles, and Eagle-2 \cite{eagle2}, the state-of-the-art for speculative decoding methods.
Additionally, we provide comprehensive comparison to these method, as well as the current fastest-to-date speculative decoding approaches, detailing the improvements SpecDiff provides in reduction of FLOPs and memory footprint in Table \ref{tab:baseline-comparison}.

\renewcommand{\arraystretch}{1.5} %
\begin{table}[t!]
    \begin{minipage}[t!]{\linewidth}
        \centering
        \resizebox{\textwidth}{!}{
        \begin{tabular}{|l|lllccr|}
            \cline{1-7}
            & & $M_p$ & $M_q$ & $\gamma$ & $\alpha$ & Speedup \\
            \cline{1-7}
            \multirow{4}{*}{\rotatebox{90}{\footnotesize \textbf{CNN/DM} }}
            & \multirow{2}{*}{\rotatebox{90}{\footnotesize \textbf{SpS} }}
              & GPT-2 XL & GPT-2  & 8 & 0.92 & 3.58x \\
            & & GPT-NEO  & GPT-2  & 9 & 0.95 & 5.45x \\
            \cline{2-7}

            &\multirow{2}{*}{\rotatebox{90}{\footnotesize \textbf{Ours} }}
            & GPT-2 XL & MDLM  & 15 & 0.87 & {\bf 4.80x} \\
            & & GPT NEO & MDLM  & 15 & 0.88 & {\bf 6.63x} \\
    

            \cline{1-7}

            \multirow{4}{*}{\rotatebox{90}{\footnotesize \textbf{OpenWebText} }}
            & \multirow{2}{*}{\rotatebox{90}{\footnotesize \textbf{SpS} }}
                & GPT-2 XL & GPT-2  & 8 & 0.93 & 3.66x \\
            & & GPT-NEO  & GPT-2  & 7 & 0.85 & 4.12x \\
            \cline{2-7}


            &\multirow{2}{*}{\rotatebox{90}{\footnotesize \textbf{Ours} }}
            & GPT-2 XL & MDLM   & 15 & 0.89 & {\bf 5.38x} \\
            & & GPT NEO & MDLM & 20 & 0.88 & {\bf 7.23x} \\
    
            
            \cline{1-7}
        \end{tabular}
        }
        \vspace{7pt}
        \caption{Evaluation of walltime speedup over autoregressive decoding using SpecDiff (Ours) compared to standard speculative decoding (SpS). The best result for each setting and target model is displayed in {\bf bold}.}
        \label{tab:walltime-metrics-prefix}
    \end{minipage}
\end{table}

\begin{table*}[t!]
\begin{minipage}[t!]{\linewidth}
\centering
    \resizebox{\textwidth}{!}{
        \begin{tabular}{|l|lllccccc|}
            \cline{1-9}
            
            \multirow{4}{*}{\rotatebox{90}{\footnotesize \textbf{MT Bench} }} & & $M_p$ & $M_q$ & $\gamma$ & $\alpha$ & FLOPs/draft  & Task Speedup & Overall Speedup \\
            \cline{2-9}

            & \multirow{1}{*}{{\footnotesize \textbf{EAGLE$^\dagger$} }}
             & Vicuna 33B  & AR Head (990M)  & $\approx$5 & 0.80 & $2.01 \times 10^{10}$ & 2.73x & 2.41x \\

            & \multirow{1}{*}{{\footnotesize \textbf{EAGLE-2$^\dagger$} }}
             & Vicuna 33B  & AR Head (990M)  & $\approx$6 &  0.84 & $8.35 \times 10^{10}$ & {\bf 3.03x} & 2.60x  \\
            \cline{2-9}

            
            & \multirow{1}{*}{{\footnotesize \textbf{Ours} }}
             & Vicuna 33B & MDLM (141M)  & 15 & 0.76 &  $5.53 \times 10^{10}$ & 2.61x & {\bf 2.61x}  \\
            \cline{1-9}
            
        \end{tabular}
        }
        \vspace{7pt}
        \caption{Evaluation on MT Bench using SpecDiff (Ours), EAGLE, EAGLE-2, and SpS with Vicuna target models. FLOPs/step computes that floating point operations for the drafter to generate each sequence. `Overall Speedups' for the baselines are measured on all {\it Spec-Bench} \cite{xia-etal-2024-unlocking} tasks.  $^\dagger$ denotes results collected using {\it Spec-Bench}.
        }
        \label{tab:walltime-llama}
    \end{minipage}
\end{table*}

\begin{figure*}[t]
    \centering
    \includegraphics[width=0.49\linewidth]{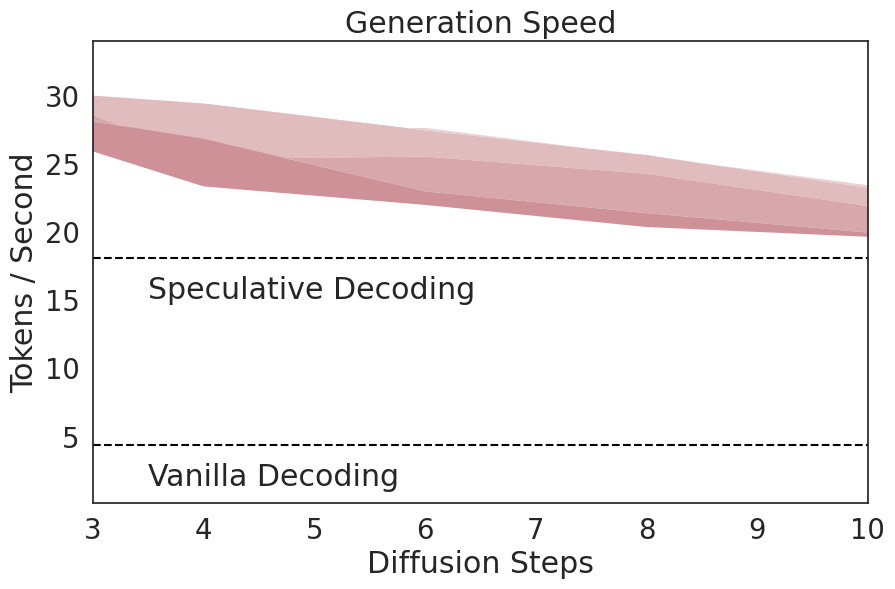}
    \includegraphics[width=0.47\linewidth]{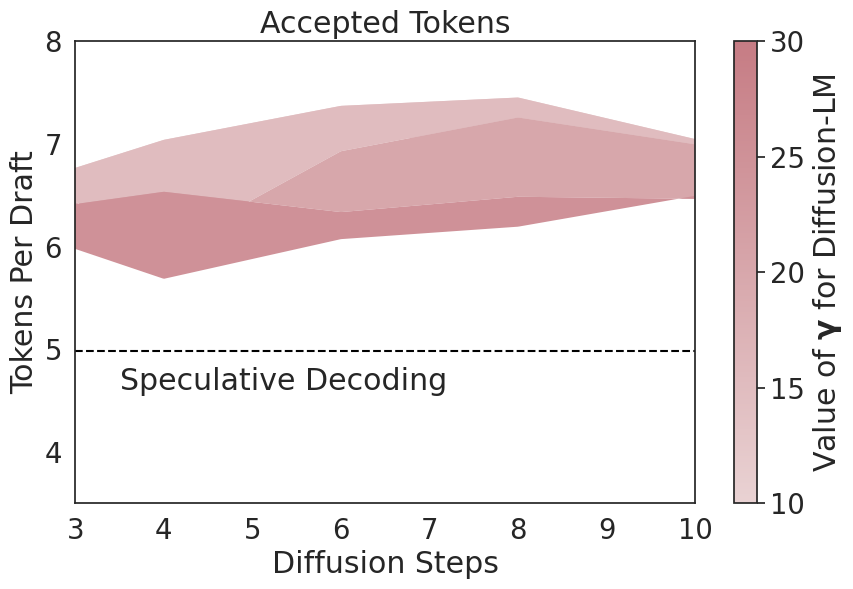}
    \caption{Evaluation of SpecDiff's sensitivity to $\gamma$ and number of diffusion steps when optimizing speed (left) and accepted tokens per draft (right) as reported on the OpenWebText task using GPT-2 NEO as the target model. Average token acceptance increases with $T$ (x-axis), as examined theoretically in Appendix \ref{appendix:theory}. Despite this, the additional compute required as $T$ is scaled results in reduced speedup.}
    \label{fig:hyperparam-analysis}
\end{figure*}

\subsection{Results and Discussion}
\label{subsec:results-discussion}

Empirically we highlight the comparison between our approach, 
other speculative decoding methods, and vanilla autoregressive generation. 
Across the tested settings and target model architectures, SpecDiff significantly outperforms standard speculative decoding, achieving speedups of up to 7.2x compared to the target models and {\bf increasing the efficiency of standard speculative decoding by more than 1.75x}. Furthermore, SpecDiff rivals the performance of current state-of-the-art methods while {\bf reducing the FLOPs/draft by over $\mathbf{33\%}$ and reducing memory consumption (shown in Table \ref{tab:baseline-comparison}).}

\paragraph{Table \ref{tab:walltime-metrics-prefix}}
First, we compare SpecDiff to the performance of standard speculative decoding. 
Despite SpS generally reporting slightly higher acceptance rates, the {\bf improved parallelization provided by SpecDiff results in greater speedups ranging from $\mathbf{1.45\mbox{--}1.75\times}$ improvement over the baseline.}
As we have not finetuned the target model or draft model, we utilize sequence initialization with SpS to enhance generation speed by sampling the first 100 tokens of the sequence.
We ablate the performance of SpecDiff without this enhancement in the Table \ref{tab:walltime-metrics}.



\paragraph{Table \ref{tab:walltime-llama}}
To provide comparison to other state-of-the-art speculative decoding methods, evaluation is conducted on generation prompts from MT Bench \cite{zheng2023judging} using Vicuna 33B as the target model.
Generation lengths are reduced to 512 tokens to better align with the task.
For this setting, MDLM (141M) is pretrained with the Llama 1 tokenizer for compatibility with the target model used.
As Vicuna is instruction tuned, we utilize a teacher-student instruction tuning approach resembling the knowledge distillation approach proposed by \citeauthor{zhou2024distillspecimprovingspeculativedecoding} to align the output distribution of MDLM with the target model.
This approach is similar to the additional instruction tuning conducted to align Eagle and Eagle-2 on ShareGPT instruction tuning dataset \cite{eagle,eagle2}.
Because of this finetuning step, it is not necessary to use sequence initialization as with pretrained models.
Notably, SpecDiff provides a speedup $70\mbox{--}85\%$ of the state-of-the-art methods, Eagle and Eagle-2, while {\bf requiring $\mathbf{33\%}$ fewer floating point operations per draft generation.} Thorough comparison to Eagle-2 and other speculative decoding strategies is highlighted in Table \ref{tab:baseline-comparison}. 


\begin{table}[!t]
    \begin{minipage}[h]{\linewidth}
        \centering
        \resizebox{\textwidth}{!}{
        \begin{tabular}{|l|lllccr|}
            \cline{1-7}
            & & $M_p$ & $M_q$ & $\gamma$ & $\alpha$ & Speedup \\
            \cline{1-7}
            \multirow{2}{*}{\rotatebox{90}{\footnotesize \textbf{CNN} }}
            
            &\multirow{2}{*}{\rotatebox{90}{\footnotesize \textbf{Ours} }}
            & GPT-2 XL & MDLM  & 10 & 0.75 & 3.82x \\
            & & GPT NEO & MDLM  & 15 & 0.77 & 5.32x \\

            \cline{1-7}

            \multirow{2}{*}{\rotatebox{90}{\footnotesize \textbf{OWT} }}

            &\multirow{2}{*}{\rotatebox{90}{\footnotesize \textbf{Ours} }}
            & GPT-2 XL & MDLM  & 15 & 0.78 & 4.06x \\
            & & GPT NEO & MDLM  & 15 & 0.82 & 6.19x \\
            \cline{2-7}
            \cline{1-7}
        \end{tabular}
        }
        \vspace{7pt}
        \caption{Performance of a ``vanilla'' version of SpecDiff, that does not use SpS initialization.}
        \label{tab:walltime-metrics}
    \end{minipage}
\end{table}

\paragraph{Table \ref{tab:walltime-metrics}}
While we observe a slight decrease in SpecDiff's speed when standard speculative decoding is not used to initialize the generation, the change in $\alpha$ is most noticeable when comparing to Table \ref{tab:walltime-metrics-prefix}. This reflects the poor acceptance rates at the beginning of the generation. While SpecDiff does outperform SpS in this setting, {\it providing up to 1.5x speedups over SpS,}  this highlights the benefit of adjointly using SpecDiff and other speculative decoding methods. Such hybrid approaches can be particularly effective for shorter generations.

While previous implementations of speculative decoding rely on a common architecture between the drafter and target models \cite{leviathan2023fast,chen2023accelerating}, using smaller versions of the same architecture to generate draft sequences, these experiments demonstrate a robustness to using a completely different architecture for sequence drafting. {\it This is particularly significant given the absence of finetuning in the reported results (Tables \ref{tab:walltime-metrics-prefix} and \ref{tab:walltime-metrics}).} Pretrained diffusion models can be directly purposed as draft models requiring no additional training.

The much larger values of $\gamma$ used for SpecDiff should particularly be highlighted. This is a key discrepancy between diffusion language models, which generate entire sequences in parallel, and autoregressive models.
Hence, there is minimal overhead to increasing the sequence length generated by the diffusion-based drafter, and $\gamma$ can be significantly increased without incurring significant cost.


The hyperparameters used in the reported results have been optimized empirically. We highlight that while in standard speculative diffusion the performance is highly sensitive to $\gamma$, SpecDiff is robust to a range of values for $\gamma$ making it unnecessary to precisely tune this hyperparameter (in our experiments we found between 10 and 20 worked well). Rather, SpecDiff's performance is much more sensitive to the number of diffusion steps selected. Similar to the role of $\gamma$ in an autoregressive model, the number of diffusion steps $T$ dictates the number of network evaluations during a single drafting step. As reported in the Figure \ref{fig:hyperparam-analysis}, while increasing this hyperparameter arbitrarily results in higher values of $\alpha$, SpecDiff performs best when this is optimized to balance the objectives of maximizing the number of accepted tokens and minimizing the drafter's overhead.

\section{Conclusion}
Motivated by the costly inference time of current large language models, this paper has proposed the novel integration of discrete diffusion models with autoregressive language models. The proposed method, Speculative Diffusion Decoding, alters existing speculative decoding schemes to integrate a non-autoregressive diffusion model as the draft model. As shown by the empirical evaluation on standard language generation benchmarks, the proposed method leverages the dramatic runtime advantages of discrete diffusion models while also maintaining the dramatically higher generation quality of autoregressive target models. The reported results demonstrate the utility of this approach in effectively accelerating runtime, outperforming vanilla decoding by over 7x and speculative decoding methods by over 1.75x.

\section{Limitations}
\label{sec:limitations}
While our proposed Speculative Diffusion Decoding method represents a significant step forward in accelerating large language model inference, several limitations warrant discussion. Addressing these limitations highlights promising avenues for future research that could further enhance the efficiency and applicability of SpecDiff.

\paragraph{Calibration of discrete diffusion models.}
A primary limitation of our approach lies in the challenge of using different architectures for the drafter and verifier models. Specifically, the adopted discrete diffusion models do not output well-calibrated probability distributions that align with those of the target autoregressive models, particularly when the sampling temperature ($T$) is greater than zero. The diffusion models tend to produce over-confident predictions, often assigning near-certain probability to the top-1 token while assigning negligible probabilities to all other tokens. This results in deterministic sampling regardless of the temperature setting, challenging the applicability of SpecDiff in scenarios where diversity and stochasticity in generation are desired.

Thus, the development of techniques to better align the output probabilities of diffusion drafters with the target models is an key area of future work. Achieving proper calibration would enable effective use of SpecDiff at higher temperatures, as well as unlocking massive further speedups beyond what we have reported, as 
the acceptance rates are predicted to increase dramatically.

\paragraph{Limited tokenization availability.}
We note that available discrete diffusion models are based on the GPT-2 tokenizer, which would restrict one immediate compatibility with target models using different tokenization schemes. For our experiments, we indeed trained a new discrete diffusion model from scratch with a different tokenizer, a process that demanded substantial computational resources and time. All our models will be released on HugginFace and thus be directly used by the community. 


\paragraph{Performance on shorter generation tasks.}
We observe that SpecDiff exhibits optimal performance on longer sequence generations. In shorter generation tasks, the benefits of parallelization are less pronounced, and finetuning the diffusion drafter may be necessary to achieve comparable efficiency gains. Without finetuning, the drafter may not effectively capture the target model’s token distributions for shorter sequences.

An interesting outcome of these observations is that tailoring the drafter to better model shorter sequences, could improve its alignment with the target model, thereby maintaining speedups even in less extensive generation tasks. We believe that this adjustment may broaden SpecDiff’s applicability. 

\paragraph{Stochastic sampling.}
Note that our experiments are conducted with the sampling temperature set to zero, resulting in deterministic token generation. While this setting simplifies the verification process it limits the exploration of the model’s capabilities in generating diverse outputs.
In the future, we plan to exploring SpecDiff’s performance at non-zero temperatures. As highlighted earlier, addressing the calibration issue of the diffusion drafter would enable effective stochastic sampling. 

\smallskip
Despite these challenges, our work lays the foundational framework for integrating discrete diffusion models with autoregressive models in speculative decoding. Each limitation discussed highlights a specific area where further research could yield significant benefits for the performance of the proposed SpecDiff. 
We are optimistic that overcoming these challenges will not only reinforce the strengths of SpecDiff but also unlock new possibilities for accelerating language model inference.

\section{Acknowledgements}

This work is partially supported by NSF awards 2143706, 2232054, and 2242931. 
This work was performed under the auspices of the U.S. Department of Energy by Lawrence Livermore National Laboratory under Contract DE-AC52-07NA27344 and was supported by the LLNL-LDRD Program under Project No. 24-ERD-010 (LLNL-CONF-2000386). This research used resources of the National Energy Research Scientific Computing Center (NERSC), a Department of Energy Office of Science User Facility using NERSC award ASCR-ERCAP0027427. The views and conclusions provided in this paper reflect those of the authors only. 

\bibliography{references}  

\newpage
\appendix

\section{Drafter Convergence Analysis}
\label{appendix:theory}

In previous speculative decoding approaches the computational overhead of the drafting stage has been proportional to $\gamma$, as this parameter dictates the number of network evaluations during the draft phase. When using a discrete diffusion draft model, the number of network evaluations is dictated by the number of diffusion steps $T$. The following reports an analysis of the possible speedups that can be achieved by our approach, as a function of the diffusion steps $T$.

First, note that the expected number of tokens per draft can be derived from $\alpha$ and $\gamma$:
\begin{equation}
\label{eq:approx_tokens}
    \mathbb{E}(\# \textit{tokens}) = \frac{1 - \alpha^{\gamma+1}}{1 - \alpha}.
\end{equation}

In prior studies, theoretical results have focused on determining the optimal $\gamma$ to maximize the throughput of speculative decoding methods \citep{leviathan2023fast}. For SpecDiff, extending $\gamma$ introduces minimal overhead and becomes less important to consider. The number of sequential diffusion operations, $T$, instead impacts $\alpha$ as increasing this number improves the convergence of $M_q(x)$ to the learned distribution, aims to closely approximate the distribution of $M_p(x)$. Hence, an implicit dependency arises between $T$ and $\alpha$, which is reflected in Figure \ref{fig:hyperparam-analysis} (right).

First, note that the computation overhead of a single network evaluation of $M_q(x)$ and $M_p(x)$ is constant. $M_q(x)$ is scaled by the number of diffusion steps, whereas all evaluations of $M_p(x)$ are conducted in parallel. Now, consider that, provided Equation \ref{eq:approx_tokens}:
\begin{equation}
    \label{eq:full_time}
    \mathbb{E}(\# \textit{tokens}/\textit{second}) = \frac{(\frac{1 - \alpha^{\gamma+1}}{1 - \alpha})}{T c_1 + c_2}
\end{equation}
where $c_1$ is the computation overheads of a single network evaluation of $M_q(x)$ and $c_2$ is the computation overheads of a single network evaluation of $M_p(x)$. 

Next, consider that the convergence of $q(x)$ to the original data distribution, which we will denote as $\hat{q}(x)$, is proportional to the $1/T$.

    

\begin{theorem}[\cite{li2023towards}]
\label{theorem:converge}
Under standard assumptions, the convergence rate of samplers based on the probability flow Ordinary Differential Equation (ODE), converge at the rate
{\small
\begin{align*}
    TV(q_1,\hat{q}_1) \leq c_3 \frac{d^2\log^4T}{T} + c_3 \frac{d^6\log^6T}{T^2} \\
    + c_3 \sqrt{d\log^3T\epsilon_\text{score}} + c_3d&(\log T)\epsilon_\text{Jacobi}
\end{align*}}
where $d$ is the dimensions of the sample, $\epsilon_\text{score}$ is the error in the score function estimation, $\epsilon_\text{Jacobi}$ is the error in the Jacobian matrices, and universal constant $c_3 > 0$.

\end{theorem}



Given that $q_1 \approx q_0$, this result provides a practical upper bound on the distance between $q(x)$ and $\hat{q}(x)$. As Theorem \ref{theorem:converge} provides an explicit relation to $T$, this can be used to determine an upper bound on the distance between $q(x)$ and $p(x)$ for a given number of diffusion steps. By the triangle inequality:
{\small
\begin{equation}
\label{eq:triangle}
    TV(p(x), q(x)) \leq TV(p(x), \hat{q}(x)) + TV(\hat{q}(x), q(x)) 
\end{equation}
}

Now, the remaining step to find the upper bound on the distance from $q(x)$ to $p(x)$ is to determine $TV(p(x), \hat{q}(x))$. First, consider that this relation can be expressed as follows:

\begin{definition}[\cite{leviathan2023fast}]
$D_{LK}(p,q) = \sum_{x} \|p(x) - M(x)\| = \sum_{x} \|q(x) - M(x)\| \text{ where } M(x) = \frac{p(x) + q(x)}{2}$
\end{definition}

\begin{corollary}[\cite{leviathan2023fast}]
\label{cor:alpha_divergence}
$\alpha = 1 - \mathbb{E}(D_{LK}(p,q)) = \mathbb{E}(\min(p,q))$
\end{corollary}

Provided Corollary \ref{cor:alpha_divergence}, $D_{LK}(p(x), \hat{q}(x))$ can be computed empirically by setting $T$ arbitrarily high and evaluating $\alpha$; subsequently, we will refer to this distance as $c_4$. We note that the $D_{LK}(p(x), \hat{q}(x))$ captures any error in the draft model's learned distribution that is introduced in Theorem \ref{theorem:converge} as $\epsilon_\text{score}$ and $\epsilon_\text{Jacobi}$, so we will set these to zero when applying the theorem.

Note that the metric $D_{LK}$ is equivalent to the discretized total variation: 
\begin{proof}
\begin{align*}
    D_{LK}(p,q) =& \sum_{x} \|p(x) - M(x)\| \\
    =& \frac{1}{2}\sum_{x} \|p(x)-q(x)\| \\
    \approx&  \frac{1}{2} \int \|p(x) - q(x)\|dx  \\
    =&  TV(p,q)
\end{align*}
\end{proof}

For practical applications such as this, the discrete total variation is used, and for the purpose of this analysis we will consider the metrics equivalent. Now, we are ready to compute a lower bound for $\alpha$ that is dependent on $T$, applying Equation \ref{eq:triangle}:
{\small
\begin{align*}
    \alpha = 1 - \mathbb{E}(D_{LK}(p,q)) = 1 - \mathbb{E}&(TV(p,q)) \\
    \geq  1 - (c_4 + TV(\hat{q}(x), q(x))) 
\end{align*}
}

\begin{equation}
\label{eq:alpha_to_t}
    \alpha \geq  1 - (c_4 + c_3 \frac{d^2\log^4T}{T} + c_3 \frac{d^6\log^6T}{T^2}) 
\end{equation}

Equation \ref{eq:alpha_to_t} provides a lower bound on $\alpha$ that is dependant on $T$. While in practice this remains computationally intractable, given that $c_3$ is unknown, this can be approximated using a surrogate network to predict this constant; this is not dissimilar from the suggestion by \citeauthor{leviathan2023fast} to optimize runtime using such an approach to predict $\alpha$.

\begin{table*}[!h]
\centering
    \begin{minipage}[t]{0.75\linewidth}
        \centering
        \resizebox{\textwidth}{!}{
        \begin{tabular}{|l|lllccccc|}
            \cline{1-9}
            & & $M_p$ & $M_q$ & Strategy & $\gamma$ & $T$ & Gen Length & Precision \\
            \cline{1-9}
            \multirow{2}{*}{\rotatebox{90}{\footnotesize \textbf{CNN} }}
            &\multirow{2}{*}{{\footnotesize \textbf{Ours} }}
            & GPT-2 XL (1.5B) & MDLM (110M) & {\it ddpm cache} & 15 & 2 & 1024 & 32 bit \\
            & & GPT NEO (2.7B) & MDLM (110M) & {\it ddpm cache}  & 15 & 2 & 1024 & 32 bit \\
            \cline{1-9}
            \multirow{2}{*}{\rotatebox{90}{\footnotesize \textbf{OWT} }}
            &\multirow{2}{*}{{\footnotesize \textbf{Ours} }}
            & GPT-2 XL (1.5B) & MDLM (110M) & {\it ddpm cache}   & 15 & 2 & 1024 & 32 bit \\
            & & GPT NEO (2.7B) & MDLM (110M) & {\it ddpm cache}  & 20 & 2 & 1024 & 32 bit \\            
            \cline{1-9}
            \multirow{1}{*}{\rotatebox{90}{\scriptsize \textbf{MT} }}
            &\multirow{1}{*}{{\footnotesize \textbf{Ours} }}
            & Vicuna (33B) & MDLM (141M) & {\it ddpm cache}  & 15 & 2 & 512 & 16 bit \\
            \cline{1-9}
        \end{tabular}
        }
        \vspace{7pt}
        \caption{Additional details on parametric setups for reported results.}
        \label{tab:hyperparams}
    \end{minipage}
\end{table*}

We are now ready to connect this to Equation \ref{eq:full_time}:
\begin{equation*}
    \label{eq:full_time_t}
    \scalebox{0.7}{$
    \mathbb{E}\left(\# \textit{tokens}/\textit{second}\right) \geq
    \frac{1 - \left(1 - c_4 - c_3 \frac{d^2\log^4T}{T} - c_3 \frac{d^6\log^6T}{T^2}\right)^{\gamma+1}}
    {c_4 + c_3 \frac{d^2\log^4T}{T} + c_3 \frac{d^6\log^6T}{T^2}} 
    \times \frac{1}{T c_1 + c_2}$}
\end{equation*}

This equation can now be solved analytically to optimize the lower bound. Practically, in the presence of a surrogate network, this can be simplified further, given convergence of the diffusion model is proportional to $1/T$.
\begin{equation}
    \label{eq:full_simple_t}
    \scalebox{0.9}{$
    \mathbb{E}(\# \textit{tokens}/\textit{second}) \geq
    \frac{1 - (1 - c_4 - c_5 \frac{1}{T})^{\gamma+1}}{c_4 + c_5 \frac{1}{T}} \times \frac{1}{T c_1 + c_2}$}
\end{equation}
Hence, the dependency between $T$ and $\alpha$ can be exploited to estimate the optimal number of diffusion steps. In our experiments we find that the optimal value of $T \leq 5$.

\section{Accepted Draft Lengths}

\begin{figure}[!ht]
    \centering
    \includegraphics[width=\linewidth]{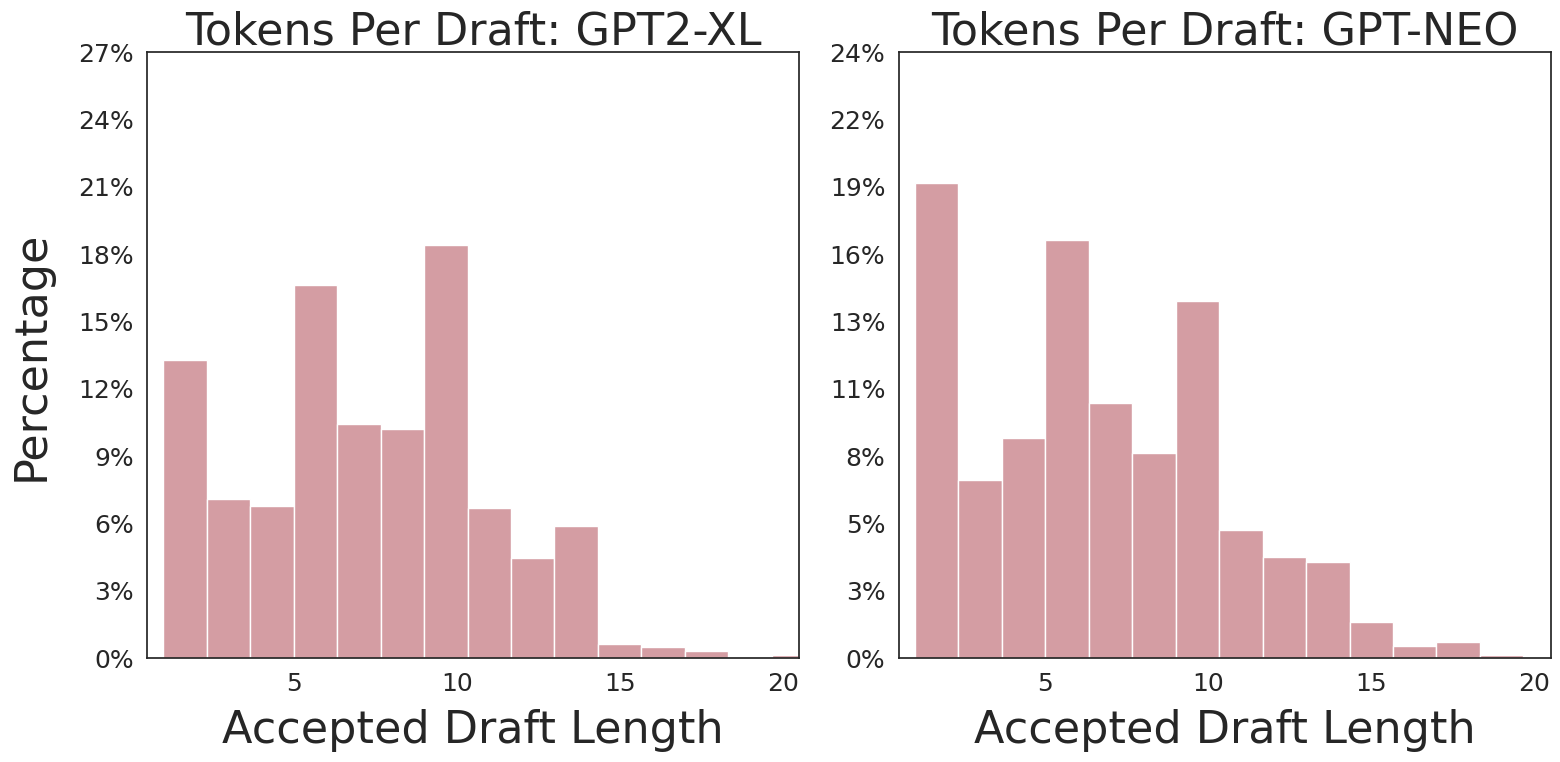}
    \caption{Accepted draft lengths for OpenWebText evaluation.}
    \label{fig:owtaccept}
\end{figure}

\begin{figure}[!ht]
    \centering
    \includegraphics[width=0.5\linewidth]{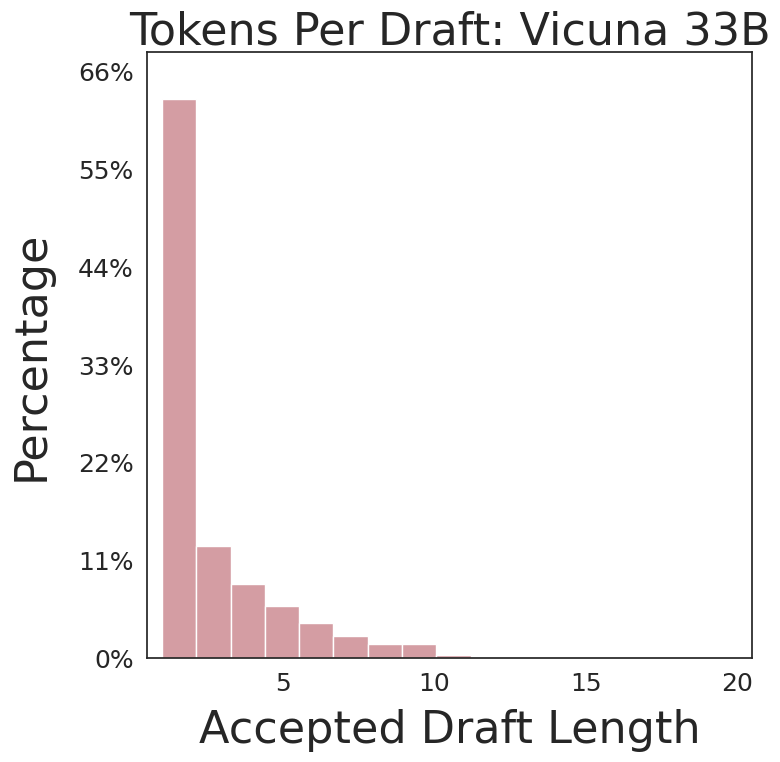}
    \caption{Accepted draft lengths for MT Bench evaluation.}
    \label{fig:mtbenchaccept}
\end{figure}

We notice a starch constrast in the lengths of accepted drafts between these two settings. While the distribution of accepted draft lengths in Figure \ref{fig:mtbenchaccept} is what would be anticipated given the lower values of $\alpha$, the longer generations on OpenWebText (Figure \ref{fig:owtaccept}) speak to the parallelism that can be realized when the distribution is effectively aligned to the target model.

\section{Implementation Details}

For all evaluation we utilize the following hyperparameter setups. If not explicitly noted here, parameters are consistent with those specified by the authors of MDLM \cite{sahoo2024simple}. 
For reported results on CNN/DM and OWT, 200 iterations are conducted on samples randomly selected from the datasets. For evaluation on MT bench, 160 iterations are conducted.

\renewcommand{\arraystretch}{1.8} 

\section{Choice of Diffusion Model}

The selection of a discrete diffusion model as the drafter plays a critical role in optimizing the overall framework’s speedup performance. The models explored, MDLM and SEDD, represent the current state-of-the-art in discrete diffusion, achieving near auto-regressive perplexity results with comparably sized models. 
We observe significant speedups over SpS when using MDLM as our drafter, as MDLM demonstrates superior generation speeds overall. These gains are not observed in SEDD for two primary reasons: first, SEDD exhibits lower perplexity compared to MDLM, resulting in a lower acceptance rate and second, MDLM’s generation speed surpasses that of SEDD.

\begin{table}[!h]
    \begin{minipage}[h]{\linewidth}
        \centering
        \resizebox{\textwidth}{!}{
        \begin{tabular}{|l|lllccr|}
            \cline{1-7}
            & & $M_p$ & $M_q$ & $\gamma$ & $\alpha$ & Speedup \\
            \cline{1-7}
            

            \cline{1-7}

            \multirow{2}{*}{\rotatebox{90}{\footnotesize \textbf{OWT} }}

            &\multirow{2}{*}{\rotatebox{90}{\footnotesize \textbf{Ours} }}
            & \multirow{1}{*}{GPT-2 XL} & SEDD Small  & 10 & 0.70 & 2.13x \\
            & & \multirow{1}{*}{GPT NEO}  & SEDD Small  & 10 & 0.77 & 2.96x \\
            \cline{2-7}
            \cline{1-7}
        \end{tabular}
        }
        \vspace{7pt}
        \caption{Performance of SpecDiff utilizing SEDD as the drafter. Note that unlike experiments in Tables \ref{tab:walltime-metrics-prefix} and \ref{tab:walltime-metrics}, the draft model has been finetuned on the selected datasets; without finetuning, acceptance rates are below $\alpha=0.2$, making SEDD impractical as a solely pretrained drafter.}
        \label{tab:sedd-metrics}
    \end{minipage}
\end{table}

\section{Temperature Impact on Acceptance Rate}
As discussed in Section \ref{sec:limitations},
overconfidence exhibited by discrete diffusion models, a result of poor model calibration, results in the top-1 token having a probability close to 1 for the vast majority of generations. 
For these outputs, the temperature is effectively zero. In the speculative decoding framework, this overconfidence is reflected as higher $q(x)$ values, which has a significant impact on the token acceptance rate when sampling stochastically. Specifically, high $q(x)$ values lead to a decrease in the number of accepted tokens because it increases the frequency of \( q(x) > p(x) \). 
Increasing the temperature has a smoothing effect on $p(x)$ leading to misaligned distributions between $p(x)$ and $q(x)$.
Consequently, tokens are more frequently rejected with a probability of \( 1 - \frac{p(x)}{q(x)} \), where a larger $q(x)$ further increasing the likelihood of rejections. 

This obstacle is avoided when sampling deterministically (temperature=0) as in the results reported. This challenge motivates future study of discrete diffusion model calibration.

\end{document}